\documentclass{article}

\usepackage{PRIMEarxiv}

\usepackage[utf8]{inputenc} 
\usepackage[T1]{fontenc}    
\usepackage{hyperref}       
\usepackage{url}            
\usepackage{booktabs}       
\usepackage{amsfonts}       
\usepackage{nicefrac}       
\usepackage{microtype}      
\usepackage{lipsum}
\usepackage{fancyhdr}       
\usepackage{graphicx}       
\graphicspath{{media/}}     
\usepackage{blindtext}
\usepackage{multirow}
\usepackage{lscape}
\usepackage{longtable}
\usepackage{makecell}
\usepackage{cite}
\usepackage{hyperref}
\usepackage{xcolor}
\usepackage{subcaption}

\newcommand{\bluebold}[1]{\textcolor{blue}{\textbf{#1}}}
\title{Consistency Evaluation of News Article Summaries Generated by Large (and Small) Language Models
}

\author{
  Colleen Gilhuly, Haleh Shahzad \\
  Royal Bank of Canada \\
  Chief Data Office \\
  Toronto\\
  \texttt{\{colleen.gilhuly, haleh.shahzad\}@rbc.com} \\
}

\begin{document}
\maketitle

\begin{abstract}
Text summarizing is a critical Natural Language Processing (NLP) task with applications ranging from information retrieval to content generation. Large Language Models (LLMs) have shown remarkable promise in generating fluent abstractive summaries but they can produce hallucinated details not grounded in the source text. Regardless of the method of generating a summary, high quality automated evaluations remain an open area of investigation. This paper embarks on an exploration of text summarization with a diverse set of techniques, including TextRank, BART, Mistral-7B-Instruct, and OpenAI GPT-3.5-Turbo. The generated summaries are evaluated using traditional metrics such as the  Recall-Oriented Understudy for Gisting Evaluation (ROUGE) Score and Bidirectional Encoder Representations from Transformers (BERT) Score, as well as LLM-powered evaluation methods that directly assess a generated summary's consistency with the source text. We introduce a meta evaluation score which directly assesses the performance of the LLM evaluation system (prompt + model). We find that that all summarization models produce consistent summaries when tested on the XL-Sum dataset, exceeding the consistency of the reference summaries.
\end{abstract}

\keywords{Text Summarization \and Large Language Models\and Natural Language Processing \and Evaluation \and Hallucination}

\section{Introduction}
\label{sec:introduction}
The rapid growth of domain-specific knowledge has
posed unprecedented challenges in efficiently utilizing the information.
With the increasing volume and diversity of such
information, generating specific and precise summaries is critical.  
An informative summary is useful as a primer and can be used in various downstream applications or for individuals who may not have time to read the entire original document. Ultimately, the goal of a summary is to enable better understanding with less time and effort.

The gold standard of summarization is human-written summaries, but there are many limitations in scalability, cost-effectiveness, and consistency. Writing a summary is a high cognitive load task and becomes much more difficult as the length of the source text increases. Some technical topics may require some subject matter expertise in addition to language proficiency, increasing the cost of the summarization task. Furthermore, human summaries are very subjective and variable which makes this approach to some extent unreliable.
Machine-generated summaries can be created on-demand and can be more easily scaled due to lower cost and time requirements. This has been an area of interest in NLP research for decades. 

Summaries can be divided into two types: extractive summaries which contain exact subsets of the source text, and abstractive summaries which contain new phrases and words that are not found in the source text.  Historically, most summary models have been extractive in nature.
Extractive summarization is considered as a sequence classification problem wherein each sentence is visited sequentially in the source document order and a binary decision is made (taking into account previous decisions made) in terms of whether or not it should be included in the summary. 

There are many examples of models trained for extractive summarization. 
For instance, \cite{nallapati2017summarunner} presents SummaRuNNer, which uses a two-layer bi-directional GRU-RNN sequence model for extractive summarization of documents.  
In \cite{durrett2016learning}, textual units are selected to include in the summary based on a rich set of sparse features whose weights are learned on a large corpus.
\cite{chopra2016abstractive} presents a conditional recurrent neural network, which acts as a decoder to generate the summary of an input sentence.

Abstractive summarization is a sequence-to-sequence problem where the output sequence (summary) generation is conditioned on the input sequence (source document). There are many instances of Transformer models of various architectures being trained for abstractive summarization. In \cite{liu2019text}, a BERT model is used for both abstractive and extractive summaries. For the extractive summary, the model acts as a classifier that predicts the sentences that should be included in the summary. To generate the abstractive summary, the problem is formulated as sequence-to-sequence problem with a standard encoder-decoder architecture. 
In \cite{hasan2021xl}, a dataset containing one million news article-summary pairs (XL-Sum) is collected and used for fine-tuning the mT5 model \cite{xue2020mt5}.
\cite{lewis2019bart} introduces BART, which learns to reconstruct the original text after it has been corrupted by a noise function. 

In the past few years, Large Language Models (LLMs) have risen to prominence as highly fluent text generators which are capable of a variety of tasks without specific task training. Instead, instructions and examples can be provided at inference time to adapt the model to the task at hand \cite{brown2020gpt3, ouyang2022instructgpt}. LLMs are able to generate coherent text with a great deal of flexibility on the style, format, and length of output, making them very well suited to a variety of summarization applications. One of the primary risks of using LLMs is their ability to generate hallucinations, which can take many forms and have varying levels of severity \cite{huang2024hallucinations}. For closed text summarization tasks, it is not desirable for the model to include general world knowledge seen during training even if it is correct and relevant to the source text. And yet, this is one of the most benign forms of hallucination. Incorrect facts or biased inferences can also appear in LLM-generated texts. 

Whether adopting an abstractive or extractive approach, it is often difficult to generate a high quality summary and even more difficult to automatically evaluate the quality of a summary in a relevant way. Criteria used in human evaluation are intangible or open-ended, and there are many ways to write a good summary. The perceived quality of a summary also depends on the context and the intended audience. Less subjective criteria like factual consistency are still difficult to evaluate automatically. Consistency remains an important metric for model-generated summaries and other texts because of LLMs' propensity for hallucinations. 

In this paper, we present an exploration of summary generation for news articles using a broad variety of both extractive and abstractive methods. We evaluate and compare the performance of these methods using a variety of techniques including standard metrics as well as LLM-powered approaches that directly evaluate consistency and hallucination. We also introduce reference-based meta evaluations which assess the overall performance of the evaluation systems (prompts and evaluator LLM). Beyond establishing correlation with human judgment, it is important to know the accuracy of LLM-powered evaluations.

\section{Data}
\label{sec:data}
We use the XL-Sum \cite{hasan2021xl} dataset as the basis for our exploration of summarization and subsequent evaluation. XL-Sum has collected one million pairs of news articles and summaries from the BBC website. The summaries are identified among the full article webpage contents by their bolded formatting. They are written by the same author as the main article and are abstractive rather than extractive. The XL-Sum dataset covers 44 languages; however English is our only language of interest for this work. There are 301k English article-summary pairs available in the dataset, with 11.5k in the test split. 

We exclude articles that are over 400 words long in order to make fair comparisons with smaller Transformer-based summarization models with a limited context window. This reduces the test split to 6484 article-summary pairs.

In XL-Sum, 37.37\% of manually reviewed English summaries were found to contain information that cannot be inferred from the article \cite{hasan2021xl}. This is a natural consequence of human authors applying their real-world background knowledge in writing the summary, but the authors note that this can also happen when the summary introduces information that is then abbreviated in the article (e.g. acronyms or referring to someone by their surname only) or for certain types of articles (e.g. blogs and opinion pieces). For a model-generated summary this would be labeled as hallucination, and we therefore wish to avoid these examples where possible. 

We note that spot-checked XL-Sum articles shorter than 100 words very frequently had some information in the summary that was not present in the article. For these short articles, the extracted summary is closer in style to a first-line summary, as in the XSum dataset \cite{narayan2018xsum}. These short articles tend to cover routine police blotter reporting or photos. 

We impose a minimum article length of 100 words in order to avoid this high concentration of summaries with lower factual overlap. 114 article-summary pairs are removed in this way, bringing our final sample size to 6370. This simple cut does not remove all instances of this problem but does offer some improvement. We will return to evaluate the factual consistency of the XL-Sum reference summaries in Section~\ref{sec:results}. No filtering or cleaning was applied to the XL-Sum test dataset other than the requirements on article length described above.

\section{Summarization methods}
\label{sec:models}
In this section, we introduce the models that we use to generate summaries in this work. Our objective is to sample a broad variety of models. The models include both extractive and abstractive approaches, and Transformer-based models of a variety of sizes. When selecting models to include, we gave some preference to model flavours that have a high number of all-time downloads on HuggingFace, therefore capturing popular models across several years. 

With the rapid improvements to existing LLMs and release of new models, the models we have tested will be behind the cutting edge of performance by the time this paper is published. This does not diminish the value of the comparison of smaller models such as T5 and BART with multi-billion parameter LLMs such as Llama and GPT-3.5-Turbo. The latest models may not be available for on-premises deployments and smaller models can shine as cost-effective solutions in many use cases. 

\subsection{TextRank}
TextRank \cite{mihalcea2004textrank} is a graph-based ranking model inspired by PageRank \cite{page1999pagerank}  that can be used in variety of NLP applications including extractive summaries. A graph is constructed where each sentence in a document is represented by a vertex and some measure of similarity represents the edges. Sentences that are highly similar to many other sentences receive a higher score and are considered more important. TextRank is better described as an algorithm than as a model, but for brevity we will collectively refer to TextRank and the models below as ``the models'' throughout this paper.

In our implementation of TextRank, we use SpaCy to split the article texts into sentences with their associated embedding vectors. Cosine similarity of the sentence embedding vectors is adopted as the similarity metric. We make use of the pagerank algorithm implementation in the python package NetworkX to produce scores. The top two sentences are taken as the TextRank summary, since summaries in the XL-Sum dataset are typically 1-2 sentences long. 

\subsection{T5-small}

The Text-to-Text Transfer Transformer (T5) model architecture offers a unified text-to-text framework where the input and output are always text strings, in contrast to BERT-style models \cite{deokar2021automated}. This allows a single model to be trained on a variety of NLP tasks using the same loss function and hyperparameters, building transfer learning into the pre-training process. 

We select a T5-small model which has been fine tuned for summarization (\url{https://huggingface.co/Falconsai/text_summarization}). The datasets used for fine-tuning are undisclosed. This model has a soft input token limit of 512 tokens, the lowest of all models covered in this work and therefore setting the maximum article length of our sample. 

 \subsection{BART-Large-CNN}
The Bidirectional and Auto-Regressive Transformers (BART) model \cite{lewis2019bart} is a sequence to sequence model named for its bidirectional (BERT-like) encoder and autoregressive (GPT-like) decoder. BART is pre-trained by corrupting text by masking tokens and shuffling sentences, and then learning a model to reconstruct the original text. The pre-training data includes 160 GB of news, books, stories, and internet text. We select a BART model fine-tuned on the CNN Daily Mail dataset (\url{https://huggingface.co/datasets/abisee/cnn_dailymail}). 

\subsection{Mistral-7B-Instruct}
Mistral-7B \cite{jiang2023mistral} is an LLM designed to balance performance and efficiency. It makes use of grouped-query attention to improve inference speed and sliding window attention to reduce the computational cost of long sequences. Mistral-7B is trained on openly available but undisclosed text sources. The base model is then fine-tuned on instruction datasets from HuggingFace and the resulting model is named Mistral-7B-Instruct. 

\subsection{Llama3-8B-Instruct}
Meta's Llama3 set of models \cite{dubey2024llama3} aim to provide a high quality set of foundation models by focusing on increased quantity and quality of training data and simplifying architecture to enable training at scale. Their cleaned and curated pre-training data includes about 15 trillion tokens in total, with a cutoff date of March 2023. The fine-tuning data includes publicly available instruction datasets, as well as over 10M human-annotated examples. We select the smallest of the Llama3 set, Llama-8B-Instruct.

\subsection{Falcon-40B-Instruct}
The Falcon series of models \cite{almazrouei2023falcon} offer open training data and detailed pre-training information in addition to open model weights in order to foster further research in the field. The training dataset, RefinedWeb \cite{penedo2023falcondata}, consists of filtered and de-duplicated web text supplemented with curated datasets. The 40B size is pre-trained on 1 trillion tokens. 

\subsection{GPT-3.5-Turbo}
OpenAI's GPT-3.5-Turbo model is a further refined version of InstructGPT \cite{ouyang2022instructgpt, brown2020gpt3} and is closely related to the first released version of the ChatGPT service which ignited the ongoing and widespread excitement for LLMs. Details around the model architecture and training data/procedure are undisclosed. We access this model via Microsoft Azure. 

\subsection{Summary prompt}
For simplicity, we use the same basic summarization prompt with minimal instructions for all of the LLMs (Mistral-7B, Llama3-8B, Falcon-40B, and GPT-3.5-Turbo): ``Write a 1-2 sentence summary of the article above.'' This may not lead to the best performance for each individual model but it does provide a shared basis for comparison with the pre-trained summarization models which do not benefit from any specialized instructions. 

The actual input to each model varies according to their chat tokenization or recommended tags to indicate instructions (i.e. \texttt{>>CONTEXT<<}, \texttt{>>QUESTION<<}, and \texttt{>>ANSWER<<} for Falcon-40B; \texttt{[INST]} and \texttt{[\textbackslash INST]} for Mistral-7B). 

All LLM summaries and evaluations are run with temperature set to zero.

\section{Performance Evaluation methods}
In this section we introduce the methods used to evaluate the model-generated summaries. The evaluation methods include standard similarity-based metrics and LLM-as-evaluator metrics. 

The standard metrics have been fading in popularity as the field of machine text summarization has advanced and as it has become clearer that these metrics have low correlation with human judgment \cite{schluter2017rouge, goodrich2019factual, kryscinski2019summarization, wang2020qags, laban2022summac}. LLM metrics offer a faster and cheaper alternative to human evaluations that can be designed to target specific criteria. 

Some LLM evaluations involve prompting the evaluator model to directly output a numerical rating for subjective but meaningful properties (such as coherence and fluency) that contribute to the perceived quality of a summary \cite{liu2023GEval, jain2023fewshot, gao2023chatgptevals}. These evaluations can also be structured as choosing the best of two candidates for a given property \cite{shen2023LLMmevallimitations, gao2023chatgptevals}. While these approaches are promising, LLMs have been demonstrated to be inconsistent evaluators, exhibiting differing degrees of alignment with human evaluations depending on the candidate or property being assessed \cite{shen2023LLMmevallimitations}. LLMs are also subject to various biases when acting as evaluators \cite{koo2024bias}. The current state of LLM evaluations can distinguish between good and bad summaries but not which of the good summaries is the best \cite{vanschaik2024fieldguide, shen2023LLMmevallimitations}.

Evaluations of factual consistency are complimentary to the above metrics which are focused on the perceived quality of a summary. In this work we implement two structured consistency evaluations which are quantitative without asking the model to directly generate a number. 

\subsection{Standard Metrics}
Standard approaches to evaluating generated text are primarily based on counting n-gram overlap. These methods assume access to one or more reference texts and score a generated text based on the precision and recall of all reference n-grams.

ROUGE-N \cite{lin2004rouge} measures the proportion of N-grams that occur in both the generated summary and the reference summary. We use $N=1$, ROUGE-1, which operates on single words. ROUGE-L, another commonly used variant, is related to the length of the longest common sub-sequence between the generated and reference summaries. The words in the sub-sequence do not need to be consecutive but must appear in the same order. For both ROUGE metrics, we report the F1 score. 

Because the ROUGE metrics are dependent on exact word choice in the generated summary, a generated summary conveying the same information with different wording would not score highly. This word-specific weakness is alleviated with BERTScore \cite{zhang2019bertscore}, which compares the semantic similarity of words by leveraging an embedding model as opposed to relying on exact matches.

Both metrics are unable to detect factual inconsistencies or model hallucinations, which are among the greatest concerns with LLM-generated summaries. These metrics also rely on one or more reference texts for comparison, and it can be expensive and challenging to obtain high-quality references summaries.  

\subsection{Question-Answer Evaluation}

\begin{figure}
    \centering
    \includegraphics[width=0.8\linewidth]{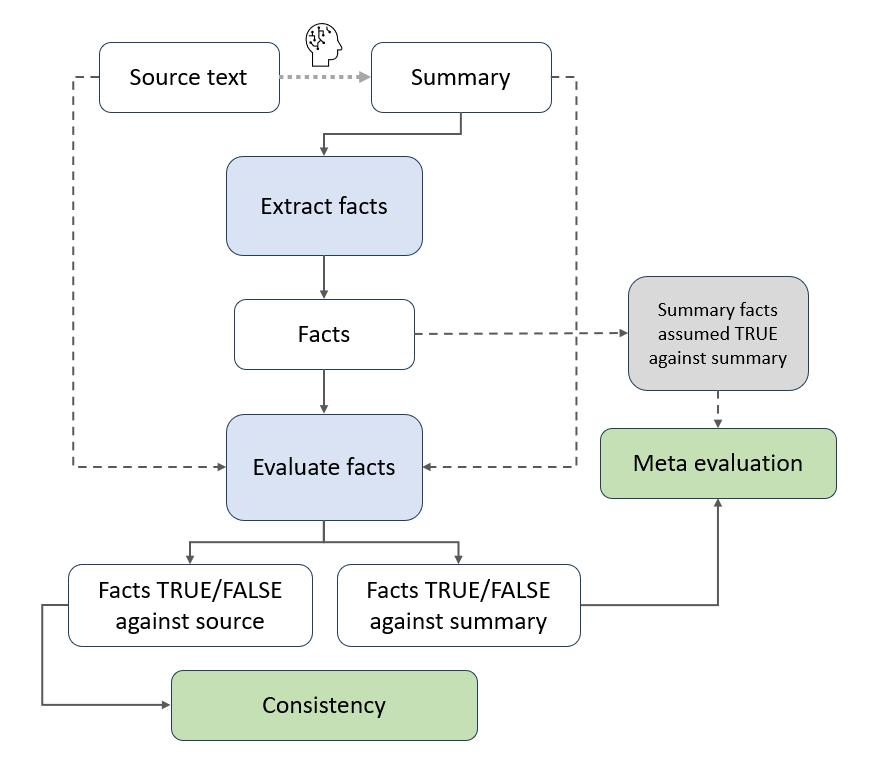}
    \caption{Overview of the LLM-powered QA evaluation.}
    \label{fig:qa_overview}
\end{figure}

Our first LLM-powered consistency evaluation is the question-answer (QA) evaluation. Earlier works relied on task-specific models for question generation and question answering \cite{chen2018QA, maynez2020faithfulness, wang2020qags, fabbri2022qafacteval, manakul2023mqag}. Challenges of this approach included filtering out low quality questions and measuring similarity of generated answers. The advent of LLMs has made fluent question generation and answering much easier to implement \cite{liu2024summequal}, though the risk of hallucinations during these tasks remains.

Whether using task-specific models or an LLM evaluator as in this work, QA evaluations present a promising alternative to standard evaluation metrics. 
The method is based on the intuition that if we ask questions about a summary and its source, we will receive similar answers if the information in the summary is factually consistent with the source.
This method can be implemented in production settings to pinpoint, and even correct, factual discrepancies in real time with minimal human oversight. 
Questions can be derived from either the summary or the source text, eliminating the need for a human-annotated reference summary. By generating questions from the source text, the informativeness of the summary can be assessed \cite{manakul2023mqag}. Conversely, by generating questions from the summary, consistency and hallucinated can be assessed. 

An overview of our QA evaluation is shown in Figure~\ref{fig:qa_overview}. Our evaluation focuses on binary (yes/no) questions generated from the summaries. Binary questions simplify the process of comparing answers from summary and source, rather than requiring another LLM call to determine whether the answers are consistent with each other (given the summary as context). We use GPT-3.5-Turbo to generate questions and answers as our evaluator LLM. 

The prompt template that we use for question generation is shown in Appendix~\ref{app:prompts}. This prompt was developed over many iterations. The first challenge was a tendency to generate only ``Yes'' questions. We therefore prompt the model to generate an answer key alongside the questions to allow later questions to be conditioned on the answers of previously generated ones. 

The question generation prompt uses generic placeholder questions to demonstrate the desired output format because we found that any specific example questions were frequently copied exactly in the output, even when the subject was not relevant to the summary at hand. Adherence to the output format is critical for clean extraction of both questions and answers for the next phase of the evaluation. 

As the summaries in our sample were generated to match XL-Sum lengths, they are short and sometimes do not have much information to create question. We include some flexibility in the prompt on the number of questions and the ability to specify an ``Unknown'' answer to avoid generating a fourth question with a verbose explanation that it cannot be answered instead of an answer key.

Regular expressions are used to extract the questions and answers from the model response. The prompt to generate question responses based on an input text (either the summary or source article) is shown in Appendix~\ref{app:prompts}. 

The answers from each response are extracted and compared against one another. The possible scenarios of the answer comparisons are shown in Table~\ref{tab:qa_scenarios}. We define the consistency score as the fraction of of questions where the summary answer equals the source answer, including when both answers are ``Unknown.'' The hallucination score is defined as the fraction of questions where the source answer equals ``Unknown'' and the summary answer does not equal ``Unknown''. Questions where the summary answer does not equal the source answer are not considered in the hallucination score. Larger values of consistency scores and lower values of hallucination scores are indicative of a highly factually accurate summary. 

We supplement these metrics with a meta evaluation of the QA evaluation system, defined as the fraction of questions where the summary answer equals the answer key that was generated alongside the question. The meta evaluation score is set to zero if some questions are unanswered or if the answer key format is incorrect. If the question generation and question answering performance of the prompt \& LLM evaluator is perfect, these answers will always be the same and the meta evaluation score would be equal to 1. 

\begin{table}
    \centering
    \begin{tabular}{|c|c|c|}
        \hline
        \multirow{2}{*}{Source Answer} &  \multirow{2}{*}{Summary Answer} & \multirow{2}{*}{Classification} \\ 
         &  & \\ \hline \hline
        Yes & Yes & Factually consistent (strong)\\ \hline
        No & No & Factually consistent (strong)\\ \hline
        Unknown & Unknown & Factually consistent (weak)\\ \hline
        Yes & No & Factual inconsistency\\ \hline
        No & Yes & Factual inconsistency \\ \hline
        Unknown & Yes/No & Hallucination\\ \hline
        Yes/No & Unknown & Non-informative \\ \hline
    \end{tabular}
    \caption{Possible scenarios of answers from source and summary}
    \label{tab:qa_scenarios}
\end{table}

\subsection{Fact-checking Method}

\begin{figure}
    \centering
    \includegraphics[width=0.8\linewidth]{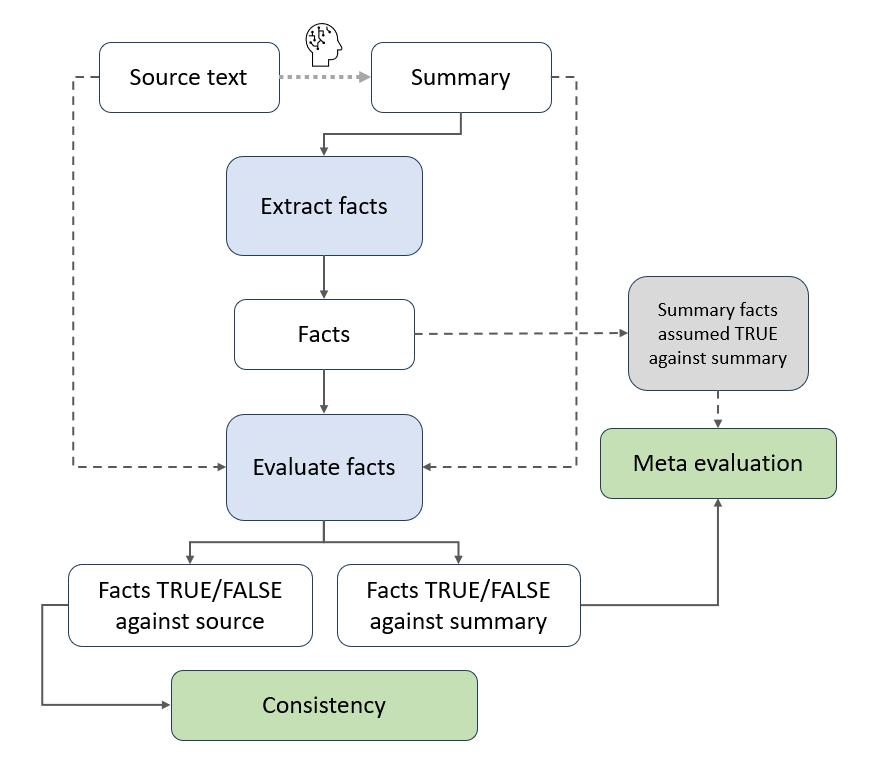}
    \caption{Overview of the LLM-powered fact-checking evaluation.}
    \label{fig:fact_overview}
\end{figure}

Our second LLM-powered consistency evaluation is a fact-checking evaluation. As with the QA evaluation, this approach was first explored and implemented with task-specific models for extracting structured fact tuples \cite{goodrich2019factual} or textual entailment \cite{maynez2020faithfulness, laban2022summac}. There has been a great deal of effort to expand upon this approach using LLMs \cite{luo2023inconsistency, tam2023factual, xu2024taxonomy, cui2024dcr}. These metrics achieve improved correlation with human evaluations, although the LLM evaluator often has positive bias towards its own generations.

Fact-checking evaluations aim to break the information contained in a summary into small, self-contained facts that can then be individually compared against the source text. This method more directly addresses consistency than the QA evaluation approach and shares the advantages of granularity, enabling automated monitoring and correcting, and not depending on a reference summary. 

An overview of our fact-checking evaluation is shown in Figure~\ref{fig:fact_overview}. Our method uses GPT-3.5-Turbo to extract a numbered list of facts from the summary and then judge whether or not they are consistent with either the summary or source article. The prompts used for these steps are shown in Appendix~\ref{app:prompts}. Notably, GPT-3.5-Turbo performed very well on the fact extraction task and very little prompt engineering was necessary for this stage. The prompt used for fact checking required more experimentation to produce answers in a consistent format without occasionally generating verbose explanations for each response. 

The true/false judgments are extracted from the model responses using regular expressions. The consistency score is defined as the fraction of summary facts determined to be true based on the source article. The meta evaluation score is defined as the fraction of summary facts determined to be true based on the summary text. If the fact extraction and evaluation performance of the prompt \& LLM evaluator is perfect, the facts will all be re-evaluated as true and the meta evaluation score would be equal to 1.

\subsection{Applying evaluations to XL-Sum reference summaries}

This work required a summarization dataset with reference summaries in order to compare standard summary evaluation metrics to LLM-powered evaluations, but the LLM-powered evaluations themselves do not require a reference summary. As an additional exercise, we apply our LLM-powered evaluations to the XL-Sum reference summaries as if they were model-generated summaries. 
It is known that a significant fraction of the reference summaries contain information not found in the corresponding articles across all languages covered in the dataset \cite{hasan2021xl}. By applying our consistency evaluations to the reference summaries, we can independently assess the frequency of unsupported information within our subset of interest. 

For completeness, we also tabulate modified ROUGE and BERTScore evaluations where the source article is treated as the reference text so that we can apply standard evaluations to the XL-Sum reference summaries. This modification significantly changes the meaning of the evaluations, particularly for the token-based ROUGE scores. 

\section{Results}
\label{sec:results}

\subsection{Summary generation}
An overview of the summarization results is shown in Table~\ref{tab:summary_stats} and example summaries are shown in Table~\ref{tab:summary_examples}. All models generate much longer summaries than the XL-Sum summaries on average, by a factor of 2 or more in word count. Llama3-8B and Mistral-7B tend to stick to the 2 sentence limit specified in the prompt but have many clauses in their sentences. Falcon-40B summaries often had more than the 2 sentence limit. 
Most models were able to successfully generate summaries for all articles in our sample. TextRank failed in 3 cases where the algorithm failed to converge. The Azure OpenAI content filter rejected 388 of the articles in our sample, which therefore could not be summarized by GPT-3.5-Turbo. These tended to be articles related to violent crimes. 

\begin{table}
    \centering
    \begin{tabular}{|c|c|c|}
        \hline 
        \multirow{ 2}{*}{Model name} &  Number of & Average summary \\
        & failed summaries & length (words) \\ \hline \hline
        TextRank & 3 & 51.5 \\\hline
        T5-small & -- & 45.2 \\\hline
        BART-large & -- & 51.0 \\\hline
        Mistral-7B & -- & 52.5 \\\hline
        Llama3-8B & -- & 61.5 \\\hline
        Falcon-40B & -- & 66.0 \\\hline
        GPT-3.5-Turbo & 388 & 39.0 \\\hline \hline
        XL-Sum Reference & -- & 20.6 \\ \hline
    \end{tabular}
    \caption{Properties of generated summaries by model}
    \label{tab:summary_stats}
\end{table}

\begin{table*}[tp]
    \centering
    \begin{tabular}{|c|p{4.5in}|}
    \hline
    Original text & The proposal features a new forest visitor centre, a medieval-style castle and areas for events and entertainment. Nottinghamshire County Council said the privately-funded venture, which could be open by spring 2015, could bring major investment into the area. At the same time Nottingham City Council said it would be looking to develop its own overhaul of the city's castle site, centred on the outlaw. The Sherwood project would mean the existing visitor facilities and car park near Edwinstowe closing down, with the new attraction built on nearby open land ... \\ \hline \hline
    XL-Sum Reference & Sherwood Forest could get a £13m Robin Hood-themed attraction under new plans. \\ \hline \hline
    TextRank & Chair of the Castle Working Group, Ted Cantle says: ``Our ambition is to seek to transform Nottingham Castle and the surrounding Castle Quarter area into a world-class, internationally recognised tourist attraction.'' The chairman of Nottinghamshire County Council's culture committee, John Cottee, said: ``The county council is committed to raising the profile of Robin and Sherwood locally, nationally and internationally, whilst at the same time protecting ancient Sherwood Forest for future generations.'' \\\hline
    T5-small &  Nottinghamshire County Council said the privately-funded venture could bring major investment into the area . The Sherwood project would mean the existing visitor facilities and car park near Edwinstowe closing down . Robin Hood is a global brand that ``deserved to be celebrated'' \\\hline
    BART-large &  The privately-funded venture could be open by spring 2015. Nottinghamshire County Council hopes it will create up to 100 jobs. The existing visitor facilities and car park near Edwinstowe will close. Nottingham City Council is also looking to develop its own overhaul of the city's castle site. \\\hline
    Mistral-7B & Nottinghamshire County Council has proposed a new forest visitor center, a medieval-style castle, and areas for events and entertainment in Sherwood Forest. The project, called ``The Sherwood Experience,'' is expected to bring major investment into the area and create up to 100 jobs. \\\hline
    Llama3-8B & Nottinghamshire County Council has proposed a new £17.50 per person attraction, ``The Sherwood Project'', featuring a forest visitor centre, medieval-style castle, and entertainment areas, which could bring major investment to the area and create up to 100 jobs. Meanwhile, Nottingham City Council is also planning to develop its own overhaul of the city's castle site, centred on the Robin Hood legend, with plans to seek funding options to meet the multi-million pound cost. \\\hline
    Falcon-40B & The article reports on a proposal to build a new forest visitor centre and medieval-style castle in Nottinghamshire, England. The privately-funded venture is expected to bring major investment into the area and create up to 100 jobs. The project is part of a larger effort to raise the profile of Robin Hood and Sherwood Forest, and to improve the tourist experience at Nottingham Castle. \\\hline
    GPT-3.5-Turbo & Nottinghamshire County Council and Nottingham City Council are both planning to develop new visitor attractions centered around the legend of Robin Hood, with the county council proposing a new forest visitor centre and the city council planning an overhaul of the city's castle site. \\\hline   
    \end{tabular}
    \caption{Example summaries generated by each model for article id uk-england-nottinghamshire-20124450, titled ``Robin Hood £13m tourist `castle' plan unveiled.'' }
    \label{tab:summary_examples}
\end{table*}

\subsection{Evaluating summaries with standard metrics}

\begin{figure*}[tbp]
    \centering
    \includegraphics[width=\linewidth]{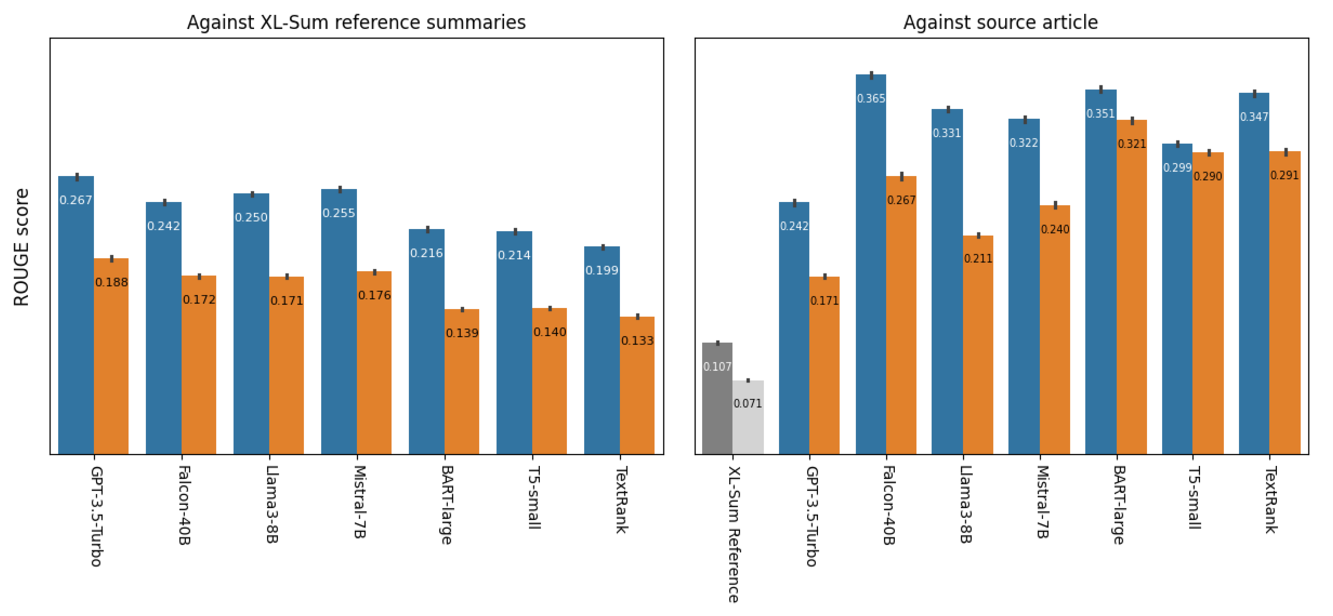}
    \caption{Average ROUGE scores for each model. The blue bars show ROUGE-1 and the orange bars show ROUGE-L. The small black error bars depict the approximate 95\% confidence interval of the averages. The left panel depicts the traditional ROUGE evaluation against reference summaries while the right panel depicts a modified ROUGE evaluation against the source article. The XL-Sum reference summaries are included in the modified evaluation on the right, depicted in grey.}
    \label{fig:rouge_results}
\end{figure*}

The average ROUGE scores for each summarization model are shown in Figure~\ref{fig:rouge_results} and Table~\ref{tab:evals}. According to both ROUGE-1 and ROUGE-L, the GPT-3.5-Turbo summaries have the greatest N-gram similarity (on average) to the XL-Sum reference summaries. This may indicate that the GPT-3.5-Turbo summaries include similar details that human authors would select, or that the language style is most similar to human authors. The other LLMs (Mistral-7B, Llama3-8B, and Falcon-40B) have similar scores, while the smaller fine-tuned models plus TextRank are a significant step lower. 

The modified ROUGE scores using the the source article text for a different perspective, and the reference summaries themselves can be evaluated in this way. The reference summaries have the lowest average ROUGE-1 and ROUGE-L by far. This is expected; human written summaries can use synonyms, apply real world knowledge, and can reflect the author's opinion. Furthermore, the reference summaries are notably shorter, which penalizes token-based similarity scores.

The modified ROUGE scores are inversely related to the degree of abstraction and condensing in the summary. GPT-3.5-turbo has the lowest score among the models considered, showing that it is most able to deviate more from simply repeating words in the article. This does not make it a ``better'' summarization model per se but certainly a more creative one for this particular task. Alternatively, this may simply reflect that the GPT-3.5-Turbo summaries are the shortest of the model-generated summaries on average. This is a metric where neither very low (zero overlap) nor very high (exact duplication) scores are desirable outcomes for summarization.  

\begin{figure*}[tbp]
    \centering
    \includegraphics[width=0.67\linewidth]{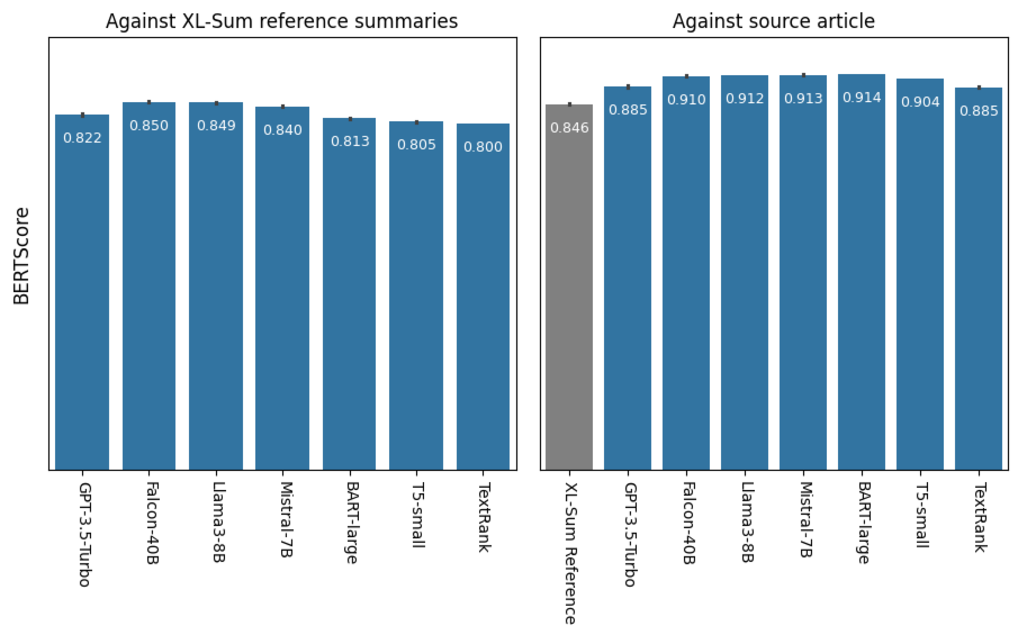}
    \caption{Average BERTScore scores for each model. The small black error bars depict the approximate 95\% confidence interval of the averages. The left panel depicts the traditional ROUGE evaluation against reference summaries while the right panel depicts a modified ROUGE evaluation against the source article. The XL-Sum reference summaries are included in the modified evaluation on the right, depicted in grey.}
    \label{fig:bertscore_results}
\end{figure*}

The average BERTScore scores for each summarization model are shown in Figure~\ref{fig:bertscore_results} and Table~\ref{tab:evals}. All models score between 0.800 and 0.850, indicating that all models generate summaries with highly semantically similar content to the reference summaries. 

When considering a modified BERTScore comparing summaries to the source articles, the XL-Sum reference summaries again receive the lowest score. However, the gap in scores is much smaller due to no longer relying on exact token matches. All model scores are slightly higher and the minimal spread among the model scores remains. 

It is interesting to see that TextRank achieves a lower or equivalent score to all other models for both types of BERTScore. The strategy of copying the two most important sentences apparently leaves out some amount semantically important information that the other summaries are able to capture. GPT-3.5-Turbo scores high for the regular BERTScore metric but low for the modified BERTScore, likely due to the shorter average summary length compared to other models. 

\subsection{LLM-powered summary evaluations}

\begin{figure*}[tbp]
    \centering
    \includegraphics[width=1\linewidth]{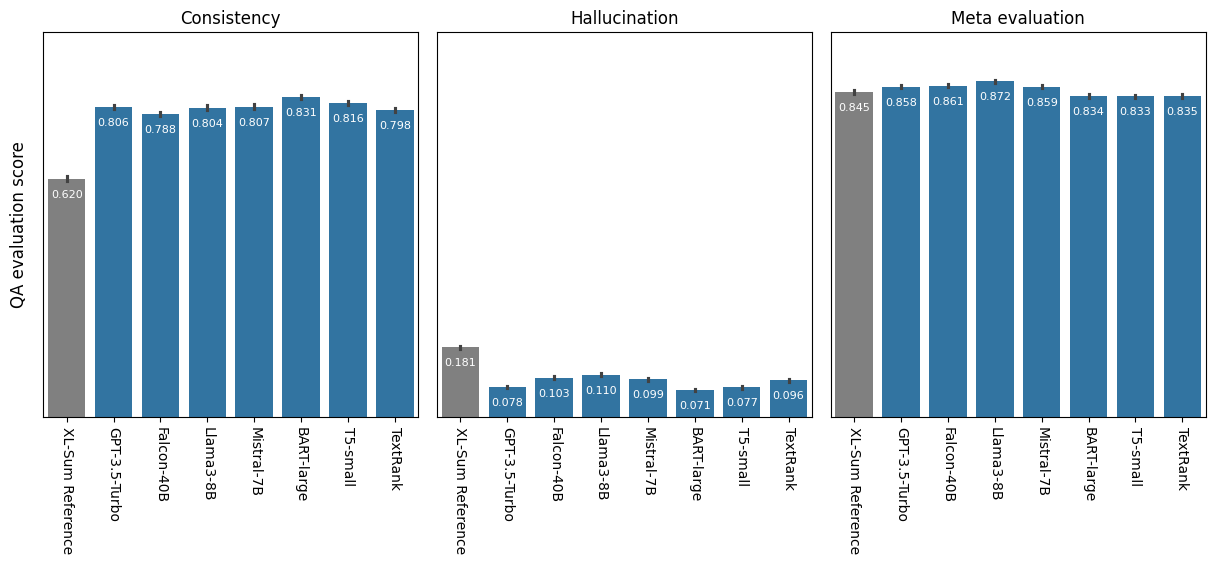}
    \caption{Average consistency, hallucination, and meta evaluation scores for the QA evaluation when applied to model generated summaries (blue bars) and the XL-Sum reference summaries (grey bar). The small black error bars depict the approximate 95\% confidence interval of the averages.}
    \label{fig:qa_results}
\end{figure*}

\begin{figure*}[tbp]
    \centering
    \includegraphics[width=0.67\linewidth]{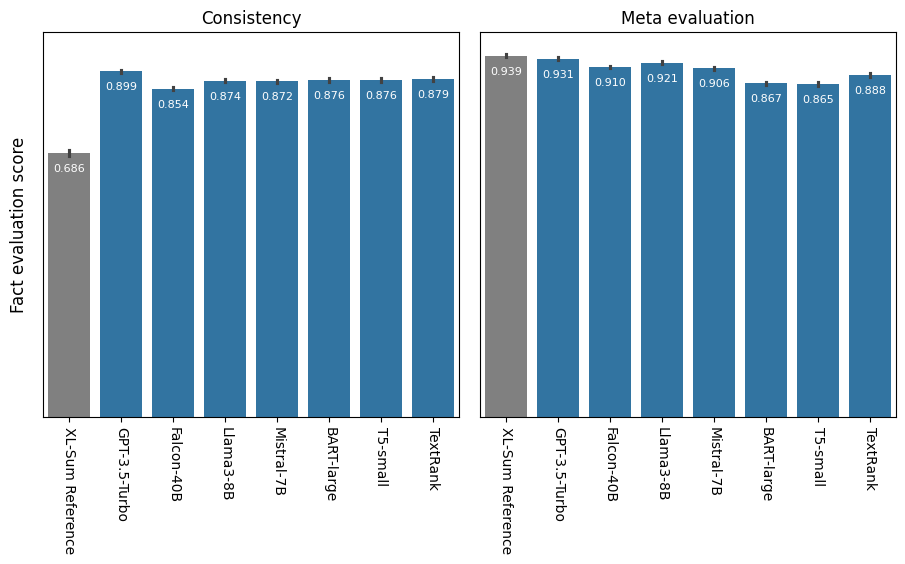}
    \caption{Average consistency and meta evaluation scores for the fact-based evaluation when applied to model generated summaries (blue bars) and the XL-Sum reference summaries (grey bar). The small black error bars depict the approximate 95\% confidence interval of the averages.}
    \label{fig:fact_results}
\end{figure*}

The results of the QA evaluation are shown in Figure~\ref{fig:qa_results} and Table~\ref{tab:evals}. There are minor differences in consistency scores across the models, all achieving roughly 80\% consistency. All model summaries have low average rates of hallucination as well. The average meta evaluation scores are significantly lower than 1.0, clearly indicating that the evaluation system is not perfectly applying the intended metrics. Due to the Azure OpenAI content filter rejecting certain harmful or violent subjects, some of our summary evaluations failed either at the question generation or question answering step. These are excluded from our results and do not negatively impact the meta evaluation scores. 

BART-large achieves the highest consistency score and the lowest hallucination score. It does not also have the highest meta evaluation score, so it is not the unambiguous highest consistency summarization model according the the QA evaluation.  

The purely extractive method TextRank is not rated as perfectly consistent nor as having zero hallucination because the evaluation system performance is not perfect. Even still, TextRank does not achieve the highest consistency or the lowest hallucination scores. It is closer to the middle or bottom of the pack in performance. 

Mistral-7B, Llama3-8B, and Falcon-40B have higher rates of hallucination than the fine-tuned summary models or GPT-3.5-Turbo. 
LLMs have greater language generation capabilities, which also comes with a greater ability to make things up. Perhaps GPT-3.5-Turbo summaries are distinguished here as the LLM with the least amount of hallucinations in this experiment, but this could also reflect an evaluator bias.

The small variations in average meta evaluation scores reflect how differences in style of the summaries can affect the accuracy of subsequent text generations. The variations in average consistency and hallucination scores among the models are much smaller than the performance gap, $1 - S_{meta}$. One could argue that it is not possible to definitively rank one model's summaries as more consistent than the other if the difference in their scores is small compared to the evaluation performance gap. 

The XL-Sum reference summaries are rated as the lowest consistency and highest hallucination rate according to the QA evaluation, with a meta evaluation score in line with that of the model-generated summaries. The difference between the XL-Sum scores and the model scores is large enough that the meta evaluation score does not cast doubt the outcome, but the exact deficit in consistency is not clear. Given the significantly lower consistency of the reference summaries, it does not provide value to evaluate the models with ROUGE or BERTScore against these references. 

The results of the fact-checking evaluation are shown in Figure~\ref{fig:fact_results} and Table~\ref{tab:evals}. All consistency and meta evaluation scores are higher than for the QA evaluation, indicating that the QA evaluation is likely underestimating consistency scores due to imperfect evaluation system performance. The fact-checking evaluation is more likely to provide an accurate relative ranking of model summary consistency due to the improved evaluation performance. As with the QA evaluation, some fact-checking evaluations failed due to the Azure OpenAI content filter and are therefore excluded from our results.

The full range of consistency scores is similar to that of the QA evaluation, but the grouping is different. Falcon-40B has the lowest consistency score (as it does for the QA evaluation), GPT-3.5-Turbo has the highest consistency score, and all of the other models are roughly halfway between these extremes. The XL-Sum reference summaries again achieve a significantly lower consistency score than the model summaries. 

Both styles of LLM evaluation have higher meta evaluation scores for the LLM summaries than those generated by pre-trained models. This could reflect a bias towards the writing style of LLMs.

\begin{landscape}
\begin{table*}[tp]
    \centering
    \begin{tabular}{|c|c|c|c|c|c|c|c||c|}
        \hline
         \multirow{2}{*}{Metric} & \multirow{2}{*}{TextRank} & \multirow{2}{*}{T5-small} & \multirow{2}{*}{BART-large} & \multirow{2}{*}{Mistral-7B} & \multirow{2}{*}{Llama3-8B} & \multirow{2}{*}{Falcon-40B} & \multirow{2}{*}{GPT-3.5-Turbo} & \multirow{2}{*}{XL-Sum Reference} \\
         & & & & & & & & \\ \hline \hline
        ROUGE-1 & {0.133} & 0.214 & 0.216 & 0.255 & 0.250 & 0.242 & \bluebold{0.267} & -- \\\hline
        ROUGE-L & {0.133} & 0.140 & 0.139 & 0.176 & 0.171 & 0.172 & \bluebold{0.188} & -- \\\hline
        ROUGE-1 (article) & 0.347 & 0.299 & 0.351 & 0.322 & 0.331 & \bluebold{0.365} & {0.242} & 0.107 \\\hline
        ROUGE-L (article) & 0.291 & 0.290 & \bluebold{0.321} & 0.240 & 0.211 & 0.267 & {0.171} & 0.071 \\\hline
        BERTScore & {0.800} & 0.805 & 0.813 & 0.840 & 0.849 & \bluebold{0.850} & 0.822 & -- \\\hline
        BERTScore (article) & {0.885} & 0.904 & \bluebold{0.914} & 0.913 & 0.912 & 0.910 & {0.885} & 0.846\\\hline
        QA Consistency & 0.798 & 0.816 & \bluebold{0.831} & 0.807 & 0.804 & {0.788} & 0.806 & 0.620 \\\hline
        QA Hallucination & 0.096 & 0.077 & \bluebold{0.071} & 0.099 & {0.110} & 0.103 & 0.078 & 0.181 \\ \hline
        Fact Consistency & 0.879 & 0.876 & 0.876 & 0.872 & 0.874 & {0.854} & \bluebold{0.899} & 0.686\\ \hline \hline
        QA Meta Evaluation & 0.835 & {0.833} & 0.834 & 0.859 & {0.872} & 0.861 & 0.858 & 0.845 \\ \hline
        Fact Meta Evaluation & 0.888 & {0.865} & 0.867 & 0.906 & 0.921 & 0.910 & {0.931} & 0.939 \\ \hline
        
    \end{tabular}
    \caption{Results of all evaluations on all model-genenerated summaries and the XL-Sum reference summaries. The highest score for each metric (or lowest score for QA Hallucination) is shown in bolded blue font.}
    \label{tab:evals}
\end{table*}
\end{landscape}

\section{Discussion}

When considering all metrics together in Table~\ref{tab:evals}, BART-large has the best or second-best scores in many cases, suggesting that it produces the most consistent summaries. GPT-3.5-Turbo also performs well, notably achieving the highest score on the LLM-powered fact-checking evaluation, but its scores may benefit from bias. Falcon-40B summaries are highly rated by standard metrics but score the lowest for both QA and fact-checking consistency, and have the second-highest QA hallucination score. The two smaller LLMs, Llama3-8B and Mistral-7B, are likely better options in terms of cost and performance when self-hosting an LLM for summary generation. 

Ultimately, all of the models produce fairly high consistency summaries. A more accurate evaluation would be required to determine which model is the most consistent (and by how much). 

\subsection{QA versus Fact-checking}

We find that our implementation of a QA-based consistency metric is less reliable than our fact-checking consistency metric, as measured by the meta evaluation scores. Both of our implementations were relatively simple and could certainly be further optimized. 

There does appear to be broader interest in fact-based \cite{luo2023inconsistency, tam2023factual, xu2024taxonomy, cui2024dcr} rather than QA evaluations \cite{liu2024summequal}. In terms of alignment with human evaluations, DCE \cite{cui2024dcr} achieves higher correlation on consistency than SummEQuAL \cite{liu2024summequal} on the SummEval benchmark when both approaches use GPT-3.5-Turbo as the evaluator. 

It is easier to prompt an LLM to extract all or nearly all of the information content of a summary than to generate questions that cover all information. The former is easier as less transformation is required on the input text. There are template-based approaches to ensure full and consistent coverage of QA evaluations, such as adopting a fixed set of questions for each type of named entity \cite{chen2018QA} or having an expert define a list of task-related questions \cite{liu2024summequal}. These approaches can generate questions that are unanswerable on both the summary and source text, but such uninformative questions could be excluded when aggregating QA responses into a score. 

\subsection{LLM capabilities as evaluators}

Previous investigations into the capabilities of LLMs as evaluators of machine-generated text have highlighted their biases \cite{koo2024bias}, inconsistent correlation with human judgment \cite{shen2023LLMmevallimitations}, and inability to identify the best of two good candidates \cite{vanschaik2024fieldguide}. With our meta evaluations, we now demonstrate that LLM evaluation systems can deviate considerably from perfect accuracy in a situation where the correct outcome is unambiguous. This shows that structured evaluations face the same challenge in differentiating candidates of similar quality, even when scores are not directly generated or head-to-head comparisons are not made. We expect higher meta evaluation scores when using a more sophisticated model as the evaluator, such as GPT-4.

A meta evaluation score that directly assesses an LLM evaluation system's accuracy provides important context to any results obtained using the system, but can also be used during development phases. For example, meta evaluation scores could be used to guide prompt engineering towards prompts that better implement the desired metric. These scores could also be used to help select an evaluator LLM. For consistency evaluations, our implementation could be extended into a two-sided approach that also evaluates the source text against itself. 

We strongly recommend the adoption of meta evaluations into a wider array of LLM evaluations where feasible. The base requirement is a reference or situation where an ideal evaluation system would always yield a perfect score. For factual consistency, this situation is comparing a text (indirectly) against itself. By definition, a text must be factually consistent with itself unless it contains explicit internal contradictions. Unfortunately, this concept does not translate well to intangible qualities related to language style. When it comes to style, there are many acceptable outcomes and it doesn't necessarily make sense to declare a given high-quality reference to be perfect.

\subsection{Evaluations applied to human benchmark summaries}

All of the consistency metrics tested in this work would agree that the XL-Sum reference summaries are the least consistent. The human authors of the XL-Sum articles and summaries had a specific objective: to catch the interest of human readers and encourage them to read the rest of the article. This is not necessarily the same as writing a consistent and comprehensive summary, and this task misalignment is important to keep in mind.

The creativity and world knowledge of the human authors is in conflict with strict factual consistency. Human authors can also include subjective interpretations based on their values in their summaries. These attributes are desirable in human-written summaries but would be classified as hallucinations in model-generated summaries. Creativity aside, there are many genuine instances in the XL-Sum dataset where the summaries contain information that is not present in the article text. To human readers, this may be desirable especially for shorter articles so that the reading experience is not too repetitive. 

There are implications for fine-tuning a summarization model on human-written summaries: do these samples provide impossible examples that effectively encourage hallucination? This is less of a concern going forward with the broad transition to foundation models versus task-specific models, but this aspect of data quality will continue to be important during pre-training and instruct fine-tuning.

The assumption underlying reference-based evaluations and training/fine-tuning on article-summary datasets is that the human written summary is the gold standard of quality and accuracy. However, all such datasets have some trade-offs between size, manual effort, and quality; prioritizing all three is rarely done due to the high cost involved. XL-Sum and other summarization datasets scraped from the internet are primarily optimizing for large size and low manual effort. 

Others have raised concerns on the quality of news summarization datasets sourced by web scraping \cite{maynez2020faithfulness, zhang2024benchmark}. We apply our LLM-powered consistency evaluations to high quality human benchmark summaries of news articles shared by \cite{zhang2024benchmark}. This also provides a test of our evaluation performance on longer articles; the summaries themselves are similar to our model-generated summaries

The average consistency scores on these human benchmark summaries are 0.761 (QA) and 0.856 (Fact-checking), both of which are approximately 0.23 points higher than the corresponding XL-Sum scores. The fact-checking consistency score on this human benchmark is now comparable with the model scores. The average QA hallucination score is 0.130, an improvement of 0.05 compared to XL-Sum. The average meta evaluation scores (0.852 and 0.915) are in line with those seen for the models and XL-Sum reference summaries, indicating that there was no substantial change in evaluation system performance on the new dataset. Overall, the high quality human benchmark summaries are significantly more factually accurate than the XL-Sum reference summaries, and the summarization models tested in this work achieve comparable factuality to this benchmark.

\subsection{Areas for future work}

The evaluation prompts and systems used in this work are relatively simple, and there are many possible avenues for improvement. For example, the evaluator LLM may produce more accurate responses if it only answers one question at a time. This would increase the time and cost of question-answering inference, which could be a problem for longer source texts. Allowing the LLM evaluator to generate ``thoughts'' about a given fact or question could also improve accuracy if it is prompted to reflect before generating its final answer \cite{wei2023cot}.

In use cases limited to a specific domain (e.g. finance, medicine), adding domain-specific context could improve both summary quality and evaluation system performance. The wide variety of topics news articles in XL-Sum dataset was not a good fit for this approach. Few-shot prompting may improve performance for specific tasks as well. We mainly experimented with few-shot prompting for question generation, and found that it led to repeating the example question even when it was not relevant to the summary. It is possible that this approach would be more applicable to a different task or a specific domain. 

In general, the investigation of strategies for LLM evaluations is ongoing. The ability to prompt in natural language leads to many creative approaches, such as having LLMs take on the roles of different stakeholders during head-to-head evaluation of two candidates \cite{wu2023roleplayer}. LLMs can be used to automatically explore design of evaluation prompts that lead to the best alignment with human evaluations \cite{liu2023autocalibrate}. Techniques such as this could also incorporate meta evaluations to test system accuracy in addition to human alignment.

\section{Conclusion}
\label{sec:conclusion}

In this paper, we evaluated the factual consistency of a wide variety of summarization models. We found that even relatively simple models such as T5-small and BART-large are able to produce highly consistent summaries. Our LLM evaluations of consistency were found to have less than ideal performance via meta evaluations of summary consistency against itself. A more accurate evaluation system would be needed to distinguish which of the summarization models produces the most consistent summaries. 

It is important to develop more reliable evaluations and continue to assess the evaluation performance not just against human preferences but also against scenarios where the correct answer is known. We recommend implementing meta evaluation scores for LLM-powered evaluations in order to guide the development process and to provide context on the final results.

Low consistency scores on the XL-Sum reference summaries indicated that this dataset is an inadequate point of comparison to modern summarization models. Likewise, traditional reference-based metrics such as ROUGE and BERTScore do not provide value when the model-generated summaries are reasonably capable of fluent writing, especially when compared to lower quality references. 

\appendix
\section{LLM Evaluation Prompts}
\label{app:prompts}

\subsection{Question Generation}

\begin{verbatim}
News article summary:
{text}

Please write 4 yes-or-no questions based on the key facts presented in the summary 
above. Keep the questions simple, with either a "Yes" or "No" answer. If the summary
is very short and you can't come up with 4 questions, 3 is also acceptable. At least
one question should have "No" as the correct answer. After each question, include
the correct answer in square brackets for the answer key, e.g.:
1. <your question here> [No]
2. <your next question here> [Yes]

Ideally, every question should be answerable using only the news article summary.
If the question cannot be answered using the summary, write "Unknown" as the
correct answer.

Questions:
\end{verbatim}

\subsection{Question Answering}

\begin{verbatim}
News article:
{text}

Based on the news article above, please answer the following numbered questions. 
Answer each question with either "Yes", "No", or "Unknown" if the answer can't be
determined based on the information in the article.

Questions:
{questions}

Answers:

\end{verbatim}

\subsection{Fact extraction}

\begin{verbatim}
{text}

Please list all of the facts presented in the summary above as a numbered list.
\end{verbatim}

\subsection{Fact checking}

\begin{verbatim}
News article:
```
{text}
```

Are the statements below factually consistent with the article? Please respond with
TRUE or FALSE accordingly for each statement, e.g.:
1. TRUE
2. TRUE
3. FALSE

----------

Statements:
{facts}

Answers (TRUE/FALSE):


\end{verbatim}

\bibliographystyle{unsrt}  
\bibliography{references}

\end{document}